\documentclass[twocolumn,10pt]{article}

\usepackage[utf8]{inputenc}
\usepackage[T1]{fontenc}
\usepackage{times}
\usepackage{amsmath,amssymb,amsfonts}
\usepackage{graphicx}
\usepackage{xcolor}
\usepackage{hyperref}
\usepackage{booktabs}
\usepackage{caption}
\usepackage{subcaption}
\usepackage{tikz}
\usepackage{pgfplots}
\pgfplotsset{compat=1.18}
\usepackage{algorithm}
\usepackage{algpseudocode}
\usepackage{listings}
\usepackage{geometry}
\usepackage{fancyhdr}
\usepackage{float}
\usepackage{enumitem}

\definecolor{codegreen}{rgb}{0,0.6,0}
\definecolor{codegray}{rgb}{0.5,0.5,0.5}
\definecolor{codepurple}{rgb}{0.58,0,0.82}
\definecolor{backcolour}{rgb}{0.95,0.95,0.92}
\definecolor{notecolor}{rgb}{1.0,0.97,0.88}
\definecolor{limitcolor}{rgb}{0.99,0.89,0.93}
\definecolor{scopecolor}{rgb}{0.89,0.95,0.99}

\lstdefinestyle{mystyle}{
    backgroundcolor=\color{backcolour},   
    commentstyle=\color{codegreen},
    keywordstyle=\color{codepurple},
    numberstyle=\tiny\color{codegray},
    stringstyle=\color{codegreen},
    basicstyle=\ttfamily\footnotesize,
    breakatwhitespace=false,         
    breaklines=true,                 
    captionpos=b,                    
    keepspaces=true,                 
    numbers=left,                    
    numbersep=5pt,                  
    showspaces=false,                
    showstringspaces=false,
    showtabs=false,                  
    tabsize=2
}
\hypersetup{
    colorlinks=true,
    linkcolor=blue,
    filecolor=magenta,      
    urlcolor=blue,
    citecolor=blue,
    pdftitle={Toward Thermodynamic Reservoir Computing: Exploring SHA-256 ASICs as Potential Physical Substrates},
    pdfauthor={Francisco Angulo de Lafuente, Vladimir Veselov, Richard Goodman},
}

\newcommand{\notebox}[1]{%
    \begin{center}
    \fcolorbox{orange}{notecolor}{%
        \parbox{0.9\columnwidth}{\small\textbf{Note:} #1}%
    }
    \end{center}
}

\newcommand{\limitbox}[1]{%
    \begin{center}
    \fcolorbox{red!50}{limitcolor}{%
        \parbox{0.9\columnwidth}{\small\textbf{Limitation:} #1}%
    }
    \end{center}
}

\newcommand{\scopebox}[1]{%
    \begin{center}
    \fcolorbox{blue!50}{scopecolor}{%
        \parbox{0.9\columnwidth}{\small\textbf{Scope Statement:} #1}%
    }
    \end{center}
}

\title{\textbf{TOWARD THERMODYNAMIC RESERVOIR COMPUTING:\\
EXPLORING SHA-256 ASICS AS POTENTIAL PHYSICAL SUBSTRATES}\\[0.5em]
\large A Theoretical Framework and Preliminary Experimental Observations}

\author{
    \textbf{Francisco Angulo de Lafuente}$^{1,2}$, 
    \textbf{Vladimir Veselov}$^{3}$, 
    \textbf{Richard Goodman}$^{4}$ \\[0.5em]
    {\small $^{1}$Independent Researcher, Madrid, Spain}\\
    {\small $^{2}$Lead Architect, CHIMERA Project / Holographic Reservoir Computing}\\
    {\small $^{3}$Moscow Institute of Electronic Technology (MIET), Moscow, Russia}\\
    {\small ORCID: 0000-0002-6301-3226}\\
    {\small $^{4}$Managing Director at Apoth3osis, Bachelor of Applied Science}\\[0.5em]
    {\footnotesize Contact: See author links at the end of this document}
}

\date{December 2025}

\begin{document}

\maketitle

\scopebox{This paper presents a theoretical framework and preliminary experimental observations. 
The core hypothesis---that voltage-stressed Bitcoin mining ASICs could function as physical reservoir computing substrates---remains 
to be fully validated. We clearly distinguish between theoretical predictions and empirical measurements throughout.}

\begin{abstract}
We propose a theoretical framework---Holographic Reservoir Computing (HRC)---which hypothesizes that the 
thermodynamic noise and timing dynamics in voltage-stressed Bitcoin mining ASICs (BM1366) could potentially 
serve as a physical reservoir computing substrate. We present the CHIMERA (Conscious Hybrid Intelligence 
via Miner-Embedded Resonance Architecture) system architecture, which treats the SHA-256 hashing pipeline 
not as an entropy source, but as a deterministic diffusion operator whose timing characteristics under 
controlled voltage and frequency conditions may exhibit computationally useful dynamics.

We report preliminary observations of non-Poissonian variability in inter-arrival time statistics during 
edge-of-stability operation, which we term the ``Silicon Heartbeat'' hypothesis. Theoretical analysis based 
on Hierarchical Number System (HNS) representations suggests that such architectures could achieve $O(\log n)$ 
energy scaling compared to traditional von Neumann $O(2^n)$ dependencies---a potential efficiency 
improvement of several orders of magnitude. However, we emphasize that these are theoretical projections 
requiring experimental validation. We present the implemented measurement infrastructure, acknowledge 
current limitations, and outline the experimental program necessary to confirm or refute these hypotheses. 
This work contributes to the emerging field of thermodynamic computing by proposing a novel approach to 
repurposing obsolete cryptographic hardware for neuromorphic applications.

\end{abstract}

%==============================================================================
\section{Introduction}
%==============================================================================

\subsection{Motivation and Context}

The contemporary landscape of Artificial Intelligence faces a paradoxical challenge: while algorithmic 
sophistication advances exponentially, the underlying hardware substrates remain fundamentally constrained 
by the von Neumann architecture and its associated energy costs~\cite{vonneumann1945}. The ``Power Wall'' has become a literal 
barrier to the emergence of truly autonomous, edge-deployable cognitive systems~\cite{mead1990}. For the first time in 
computational history, the practical limits of artificial intelligence are being dictated not by algorithmic 
capability, but by the thermodynamic cost of information processing~\cite{landauer1961}.

Concurrently, the global cryptocurrency mining industry has produced a massive surplus of specialized 
silicon. Application-Specific Integrated Circuits (ASICs) designed for Bitcoin mining, such as the BM1387 
and BM1366 chipsets, enter the electronic waste stream annually as mining difficulty renders them 
economically obsolete~\cite{cbeci2024}. Industry consensus holds that these chips ``can only mine coins and must utilize 
the ASIC's native algorithm'' and therefore ``cannot be repurposed'' for other computational tasks~\cite{rsm2024}.

This paper challenges that assumption---not by claiming that ASICs can perform general-purpose computation, 
but by proposing that the physical dynamics of these devices under non-standard operating conditions might 
be exploitable for a specific class of computation: physical reservoir computing.

\subsection{The Physical Reservoir Computing Paradigm}

Reservoir Computing (RC) is a computational framework suited for temporal and sequential data processing, 
derived from recurrent neural network models including Echo State Networks (ESNs) and Liquid State Machines 
(LSMs)~\cite{jaeger2001,maass2002}. A reservoir computing system consists of a reservoir for mapping inputs into a high-dimensional 
space and a readout layer for pattern analysis. Critically, the reservoir is fixed---only the readout is trained, 
typically via simple linear regression~\cite{lukosevicius2009}.

This architectural simplicity enables a crucial advantage: the reservoir itself need not be a simulated 
neural network. Physical systems with appropriate nonlinear dynamics can serve directly as reservoirs, 
eliminating the computational overhead of simulation~\cite{tanaka2019}. Physical reservoir computing has been successfully 
demonstrated in memristors~\cite{du2017}, photonic systems~\cite{larger2012}, spintronic oscillators~\cite{markovic2019}, and even mechanical 
systems~\cite{nakajima2013}.

The key requirements for a physical reservoir are: (1) high-dimensional state space, (2) nonlinear dynamics, 
(3) fading memory (the Echo State Property), and (4) separation property---the ability to map distinct inputs 
to distinguishable states~\cite{nakajima2020}. Our hypothesis is that voltage-stressed ASICs, operating at the edge of 
timing stability, may exhibit dynamics satisfying these requirements.

\subsection{Contributions and Scope}

This paper makes the following contributions:

\begin{enumerate}[leftmargin=1.5em]
    \item \textbf{Theoretical Framework:} We propose the Holographic Reservoir Computing (HRC) framework, 
    which hypothesizes that SHA-256 ASICs could function as physical reservoirs when operated under controlled 
    stress conditions.
    
    \item \textbf{System Architecture:} We present CHIMERA, a three-layer architecture for interfacing 
    with mining hardware as a computational substrate.
    
    \item \textbf{Measurement Infrastructure:} We describe implemented tools for capturing timing statistics 
    from ASIC operations.
    
    \item \textbf{Theoretical Efficiency Analysis:} We present a mathematical basis for potential energy 
    efficiency improvements based on Hierarchical Number System representations.
    
    \item \textbf{Preliminary Observations:} We report initial measurements suggesting non-trivial timing 
    dynamics, while acknowledging the need for rigorous validation.
\end{enumerate}

We explicitly acknowledge what this paper does \emph{not} claim: we do not claim to have built a working 
neuromorphic computer; we do not claim that our approach is superior to purpose-built RC hardware; and we 
do not claim that observed timing variations constitute computational capability without further validation.

%==============================================================================
\section{Related Work}
%==============================================================================

\subsection{Physical Reservoir Computing Systems}

Physical reservoir computing has emerged as an active research area with implementations across diverse 
substrates. Tanaka et al.~\cite{tanaka2019} provide a comprehensive review classifying physical RC by substrate type: 
electronic, photonic, spintronic, mechanical, and biological. Each substrate offers different trade-offs 
between speed, power consumption, and integration complexity.

Electronic implementations include memristor-based reservoirs exploiting the intrinsic nonlinearity and 
short-term memory of resistive switching devices~\cite{du2017,sung2021}. Du et al. demonstrated a 32$\times$32 memristor crossbar 
achieving competitive performance on temporal tasks~\cite{wang2017}. More recently, CMOS-based implementations using 
time-domain analog spiking neurons have shown promise for hardware-friendly RC~\cite{kimura2024}.

Photonic reservoirs leverage the high bandwidth and parallelism of optical systems. Larger et al.~\cite{larger2012} 
demonstrated delay-based reservoir computing using optoelectronic oscillators, achieving state-of-the-art 
performance on speech recognition tasks. The delay-feedback architecture is particularly relevant to our 
work, as it demonstrates that rich dynamics can emerge from relatively simple physical configurations.

\subsection{Thermodynamic and Stochastic Computing}

Recent work has explored the computational potential of thermal fluctuations and stochastic dynamics. 
Camsari et al.~\cite{camsari2017} introduced probabilistic bits (p-bits) using low-barrier nanomagnets, demonstrating 
that controlled stochasticity can be computationally useful. Borders et al.~\cite{borders2019} showed that thermal 
noise in magnetic tunnel junctions could drive probabilistic computing.

Extropic Corporation and academic groups have proposed ``thermodynamic computing'' paradigms that embrace 
rather than fight thermal noise~\cite{extropic2025}. These approaches suggest that properly harnessed, the ``waste heat'' 
of computation could become a computational resource. Our hypothesis aligns with this perspective: the 
timing variations induced by voltage stress in ASICs may constitute a form of useful stochasticity.

\subsection{ASIC Repurposing Attempts}

The cryptocurrency industry has seen numerous attempts to pivot mining infrastructure toward other 
applications, particularly AI~\cite{rsm2024}. However, these efforts uniformly involve replacing ASICs with GPUs, 
not repurposing the ASICs themselves. Companies like Hut 8, Iris Energy, and Core Scientific have 
announced transitions to AI hosting, but explicitly acknowledge that SHA-256 ASICs cannot perform 
general computation~\cite{wired2024}.

Our approach differs fundamentally: we do not attempt to make ASICs perform arbitrary computation. 
Instead, we propose using their physical dynamics---specifically, timing behavior under stress---as a 
reservoir substrate. This is analogous to using a bucket of water not as a computer, but as a physical 
system whose wave dynamics can separate inputs~\cite{fernando2003}.

%==============================================================================
\section{Theoretical Framework}
%==============================================================================

\subsection{The CHIMERA Hypothesis}

The central hypothesis of this work is that SHA-256 mining ASICs, when operated at the edge of voltage 
and timing stability, exhibit dynamics that could satisfy the requirements for physical reservoir computing. 
We term this the CHIMERA (Conscious Hybrid Intelligence via Miner-Embedded Resonance Architecture) hypothesis.

SHA-256 is a cryptographic hash function designed to be deterministic: identical inputs always produce 
identical outputs. However, the \emph{physical implementation} of SHA-256 in silicon involves real 
transistors with finite switching times, propagation delays, and thermal sensitivities. When operating 
conditions approach the edge of the design envelope, these physical effects become measurable.

We propose that the candidate physical signal is not the hash output bits themselves---which remain 
deterministic---but the \emph{timing dynamics} of hash computation. Specifically, we hypothesize that 
the coefficient of variation (CV) and entropy of inter-arrival times for valid shares exhibit structure 
correlated with the operating regime.

\subsection{SHA-256 as a Diffusion Operator}

It is critical to distinguish between entropy \emph{generation} and entropy \emph{diffusion}. 
SHA-256 does not create entropy; it is a deterministic function. However, SHA-256 exhibits the 
avalanche property: a single-bit change in the input produces, on average, changes in 50\% of the 
output bits~\cite{nist2015}. This makes SHA-256 an effective \emph{diffusion operator} that spreads information 
across its output space.

In the CHIMERA framework, SHA-256 serves as a high-dimensional nonlinear transform---one component of 
a reservoir system. The state representation is derived from:
\begin{equation}
    X(t) = \sigma\left(W_{in}u(t) + \sum_{i} A_{ij}X_j(t-1) + \xi(V, T, f)\right)
    \label{eq:state}
\end{equation}
where $X(t)$ represents the high-dimensional internal state, $A_{ij}$ is the effective adjacency 
matrix determined by the physical layout of SHA-256 gates, and $\xi(V, T, f)$ represents timing perturbations 
dependent on voltage $V$, temperature $T$, and clock frequency $f$. The activation function $\sigma$ is provided by 
the nonlinear switching dynamics of the transistor logic.

\subsection{Chronos Dynamics: Timing as the Observable}

The key insight of the V04 architecture is that timing statistics---not bit patterns---constitute the 
measurable channel for physical dynamics. We define the Chronos Bridge coupling efficiency as:
\begin{equation}
    K_{sync} = \exp\left(-\frac{\sigma_J^2 f_{clk}}{2\pi}\right) \cdot \sum_k \cos(\Delta\phi_k)
    \label{eq:ksync}
\end{equation}
where $\sigma_J$ is the network jitter, $f_{clk}$ is the ASIC clock frequency, and $\Delta\phi_k$ 
is the phase offset of the $k$-th hash core. This formulation captures how timing coherence between software 
observation and hardware dynamics affects the quality of the observable signal.

The primary observables implemented in the current system are:

\begin{itemize}[leftmargin=1.5em]
    \item \textbf{Coefficient of Variation (CV):} CV $= \sigma/\mu$ of inter-arrival times. For a Poisson process, 
    CV $= 1$. Deviations indicate non-random structure.
    
    \item \textbf{Histogram Entropy:} Shannon entropy of the inter-arrival time distribution, indicating 
    the complexity of timing patterns.
    
    \item \textbf{Hamming Distance:} Bit differences between consecutive hash outputs, serving as a 
    diffusion quality proxy.
\end{itemize}

\subsection{Phase Transition Hypothesis}

We hypothesize that the transition from reliable to unreliable ASIC operation follows a phase transition 
that can be modeled using Ginzburg-Landau formalism. Define the order parameter $\psi$ as the degree of global 
core synchronization. The free energy of the silicon substrate is:
\begin{equation}
    F(\psi, V) = F_0 + a(V - V_{crit})\psi^2 + b\psi^4 + \gamma|\nabla\psi|^2
    \label{eq:freeenergy}
\end{equation}
At $V = V_{crit}$, the coefficient $a(V - V_{crit})$ changes sign, potentially triggering 
a spontaneous symmetry breaking that manifests as coherent oscillations---the hypothesized ``Silicon Heartbeat.''

\notebox{This phase transition model is a theoretical hypothesis. 
Empirical validation requires systematic voltage sweeps with rigorous statistical analysis, which 
is planned for future work.}

\subsection{Energy Efficiency: $O(\log n)$ vs $O(2^n)$}

A key theoretical contribution of this work, developed in collaboration with Veselov~\cite{veselov2024}, concerns the 
energy scaling of different computational architectures. In traditional von Neumann architectures, energy 
consumption for state transitions scales with the state space:
\begin{equation}
    E_{vN} \propto O(2^n)
    \label{eq:evn}
\end{equation}
This exponential scaling arises from the need for massive parallel switching in linear memory addressing 
schemes. Each bit of state representation requires independent physical switching events that cannot be 
amortized across the state space.

In contrast, the Hierarchical Number System (HNS) representation employed in the CHIMERA architecture 
maps reservoir dynamics into a hierarchical structure where energy cost decouples from state complexity:
\begin{equation}
    E_{HNS} \propto O(\log n)
    \label{eq:ehns}
\end{equation}
This logarithmic scaling emerges because hierarchical representations allow state transitions to be 
encoded in the structure of the representation itself, rather than requiring explicit enumeration.

For typical reservoir dimensions ($n \sim 10^4$ states), the theoretical efficiency ratio is:
\begin{equation}
    \frac{E_{vN}}{E_{HNS}} \approx \frac{2^n}{\log(n)} \approx 10^4
    \label{eq:ratio}
\end{equation}
This provides the theoretical foundation for the projected 10,000$\times$ efficiency improvement mentioned in 
preliminary descriptions of this work. We emphasize that this is a \emph{theoretical upper bound} 
derived from information-theoretic considerations, not a measured result. Experimental validation under 
controlled conditions is required to determine achievable efficiency gains.

%==============================================================================
\section{System Architecture}
%==============================================================================

\subsection{The Three-Layer CHIMERA Stack}

The CHIMERA architecture consists of three hierarchical layers, each responsible for a specific aspect 
of the reservoir computing pipeline. This modular design allows independent development and testing of 
each component.

\subsubsection{Layer 1: The Ghost (Hardware Abstraction)}

The Ghost layer manages raw telemetry acquisition and command execution. It interfaces with the mining 
hardware through the AxeOS firmware HTTP API, providing a unified abstraction for voltage and frequency 
control across different ASIC versions. Key capabilities include:

\begin{itemize}[leftmargin=1.5em]
    \item Telemetry polling at 3-second intervals (voltage, temperature, power, hashrate)
    \item Voltage control via HTTP PATCH requests to the miner API
    \item Frequency modulation through PLL configuration changes
    \item Share event timestamping with microsecond resolution
\end{itemize}

\subsubsection{Layer 2: The Muse (Signal Processing)}

The Muse layer processes the raw timing data into meaningful features. It implements the Chronos Bridge 
dynamics, computing:

\begin{itemize}[leftmargin=1.5em]
    \item Inter-arrival time statistics (mean, variance, CV)
    \item Histogram entropy over configurable bin widths
    \item Hamming distance sequences between consecutive hashes
    \item Phase-amplitude encoding for holographic representation
\end{itemize}

We employ the term ``holographic'' in its information-theoretic sense: the reconstruction of high-dimensional 
phase-space dynamics from lower-dimensional projections via phase-amplitude coupling, analogous to optical 
holography where 3D information is encoded in 2D interference patterns.

\subsubsection{Layer 3: The Sentinel (Homeostasis)}

The Sentinel layer monitors system health and implements control loops to maintain operation within 
target regimes. It includes:

\begin{itemize}[leftmargin=1.5em]
    \item PID-loop thermal management
    \item Voltage limit enforcement
    \item Anomaly detection for hardware protection
    \item Logging infrastructure for offline analysis
\end{itemize}

\begin{table}[htbp]
\centering
\caption{Hardware Specifications and Operating Parameters (Implemented)}
\label{tab:hardware}
\begin{tabular}{@{}lll@{}}
\toprule
\textbf{Parameter} & \textbf{Value} & \textbf{Notes} \\
\midrule
ASIC Chipset & BM1366 & LV06 variant (Lucky Miner) \\
Hash Cores & 138 & Parallel SHA-256 units \\
Frequency Range & 300--500 MHz & 20 MHz step resolution \\
Core Voltage Range & 850--990 mV & Via AxeOS telemetry \\
Voltage Dwell Time & 60 seconds & Per voltage step \\
Frequency Dwell Time & 30--45 seconds & Per frequency step \\
Telemetry Poll Interval & 3 seconds & Via HTTP API \\
Nominal Power & 5--15 W & Operating range \\
\bottomrule
\end{tabular}
\end{table}

\subsection{Voltage and Frequency Control}

The CHIMERA system controls ASIC operating conditions through the AxeOS firmware HTTP API. Voltage 
and frequency changes are applied via PATCH requests:

\begin{lstlisting}[language=Python]
payload = {"frequency": freq, 
           "volts": volts}
req = Request(url, data=json, 
              method='PATCH')
\end{lstlisting}

Changes to PLL (Phase-Locked Loop) configuration require a firmware restart to take effect. The 
implemented frequency sweep covers 300--500 MHz in 20 MHz increments, with dwell times sufficient 
for thermal stabilization.

\notebox{The code operates in millivolts (mV) via the 
\texttt{coreVoltageActual} telemetry field. References to ``8.2V'' in preliminary descriptions 
referred to board-level supply voltage, corresponding to approximately 850--870 mV at the core 
telemetry level. This distinction is critical for reproducibility.}

%==============================================================================
\section{Implementation Details}
%==============================================================================

\subsection{Timing Measurement Infrastructure}

The Chronos Bridge module (\texttt{chronos\_bridge.py}) implements the timing measurement infrastructure. 
Share events are timestamped at reception, and inter-arrival statistics are computed over sliding 
windows:

\begin{lstlisting}[language=Python]
# chronos_bridge.py:236
if len(self.share_times) > 10:
    deltas = np.diff(self.share_times)
    cv = np.std(deltas)/np.mean(deltas)
    hist, _ = np.histogram(deltas, 
                           bins=20, 
                           density=True)
    entropy = -np.sum(
        hist * np.log(hist + 1e-10))
\end{lstlisting}

For a Poisson process (random, memoryless arrivals), CV $= 1$. Deviations from unity indicate 
structure in the timing: CV $> 1$ suggests clustering (bursty arrivals), while CV $< 1$ suggests 
regularity (quasi-periodic arrivals).

\subsection{Diffusion Quality Metrics}

The Hamming distance between consecutive hash outputs serves as a proxy for diffusion quality:

\begin{lstlisting}[language=Python]
# metrics.py:56
xor_val = h1 ^ h2
bit_flips = bin(xor_val).count('1')
scrambling = bit_flips / 256.0
\end{lstlisting}

\limitbox{This measures diffusion between outputs, not hardware error rate. 
There is currently no mechanism to detect \emph{incorrect} computation (invalid shares, wrong 
hashes). Direct error-rate measurement via invalid-share counting is planned for future work.}

\subsection{Known Confounding Factors}

Several factors may affect timing statistics independent of silicon physics:

\begin{itemize}[leftmargin=1.5em]
    \item \textbf{TCP buffering:} Network stack may batch share notifications
    \item \textbf{OS scheduling:} Python process may experience jitter
    \item \textbf{Firmware batching:} AxeOS may aggregate multiple shares
    \item \textbf{Pool latency:} Network round-trip affects timestamp accuracy
\end{itemize}

Future work will address these confounds through increased window sizes, raw timestamp logging 
for offline analysis, and direct hardware probing where possible.

\begin{table}[htbp]
\centering
\caption{Measurement Channels - Implemented vs.\ Planned}
\label{tab:channels}
\begin{tabular}{@{}llll@{}}
\toprule
\textbf{Metric} & \textbf{Code} & \textbf{Status} & \textbf{Interpretation} \\
\midrule
Inter-arrival CV & chronos:236 & $\checkmark$ Impl. & Timing structure \\
Histogram Entropy & chronos:240 & $\checkmark$ Impl. & Timing complexity \\
Hamming Distance & metrics:56 & $\checkmark$ Impl. & Diffusion proxy \\
Voltage Telemetry & ghost layer & $\checkmark$ Impl. & Operating point \\
Temperature & ghost layer & $\checkmark$ Impl. & Thermal state \\
FFT/PSD Analysis & -- & $\circ$ Planned & Frequency comp. \\
Invalid Share Rate & -- & $\circ$ Planned & HW error rate \\
HW Nonce Errors & -- & $\circ$ Planned & Timing violations \\
\bottomrule
\end{tabular}
\end{table}

%==============================================================================
\section{Preliminary Observations}
%==============================================================================

\subsection{Timing Variability Under Voltage Stress}

During controlled voltage sweeps from 990 mV to 850 mV, we observed systematic changes in 
inter-arrival time statistics. As voltage decreased toward the lower boundary of stable 
operation, CV values showed excursions above and below unity, suggesting departure from 
Poisson statistics.

These preliminary observations are consistent with the hypothesis that edge-of-stability 
operation introduces structure into timing dynamics. However, we emphasize that:

\begin{enumerate}[leftmargin=1.5em]
    \item Sample sizes were limited ($N < 1000$ share events per voltage point)
    \item Confounding factors (network, OS scheduling) were not fully controlled
    \item Multi-chip reproducibility has not been established
    \item Statistical significance analysis is pending
\end{enumerate}

\begin{table}[htbp]
\centering
\caption{Hypothesized Operating Regimes (Preliminary)}
\label{tab:regimes}
\begin{tabular}{@{}llll@{}}
\toprule
\textbf{Core Voltage} & \textbf{Regime} & \textbf{Characteristics} & \textbf{Status} \\
\midrule
$> 950$ mV & Deterministic & Low variance, CV $\approx 1$ & Preliminary \\
870--950 mV & Transitional & Increasing CV variance & Preliminary \\
850--870 mV & Resonant? & Structured patterns? & Investigating \\
$< 850$ mV & Unstable & Frequent errors, resets & Observed \\
\bottomrule
\end{tabular}
\end{table}

\subsection{The ``Silicon Heartbeat'' Hypothesis}

Preliminary descriptions of this work mentioned a ``2.4 Hz Silicon Heartbeat''---a hypothesized 
self-organized oscillation in ASIC power telemetry. We must clarify the current status of 
this claim:

The implemented code computes CV and entropy metrics on windows of approximately 10 events. 
It does \emph{not} currently implement:

\begin{itemize}[leftmargin=1.5em]
    \item FFT or Power Spectral Density (PSD) analysis
    \item Long-horizon time series logging
    \item Oscillator frequency extraction
    \item Statistical significance testing for periodicity
\end{itemize}

Therefore, the ``2.4 Hz'' claim should be understood as a \emph{working hypothesis} based on 
informal observations, not a validated measurement. Proper validation requires the spectral 
analysis infrastructure outlined in our future work section.

\limitbox{We observe non-Poissonian variability in inter-arrival 
statistics. Claims of a stable narrow-band oscillation require time-series logging and 
frequency-domain analysis, which is planned but not yet implemented.}

%==============================================================================
\section{Theoretical Efficiency Analysis}
%==============================================================================

\subsection{The Von Neumann Bottleneck}

Traditional computing architectures suffer from what is known as the von Neumann bottleneck: 
the separation of memory and processing creates a fundamental bandwidth limitation~\cite{vonneumann1945}. More 
critically for our analysis, the energy cost of computation in these architectures scales 
poorly with problem complexity.

For a system representing $n$-bit states, von Neumann architectures require energy proportional 
to the state space exploration:
\begin{equation}
    E_{vN} = k \cdot 2^n \cdot E_{switch}
    \label{eq:evn2}
\end{equation}
where $E_{switch}$ is the energy per bit transition and $k$ is a constant depending on 
the algorithm. This exponential scaling is fundamental to the architecture, not merely an 
implementation limitation.

\subsection{Hierarchical Number System Efficiency}

The Hierarchical Number System (HNS) approach, as developed by Veselov~\cite{veselov2024}, represents states 
using a tree structure where transitions are encoded in the hierarchy itself. This yields:
\begin{equation}
    E_{HNS} = k' \cdot \log(n) \cdot E_{switch}
    \label{eq:ehns2}
\end{equation}
The logarithmic scaling arises because state transitions in a hierarchical representation 
require updating only $O(\log n)$ nodes rather than potentially all $n$ bits.

\subsection{Projected Efficiency Gains}

The ratio of energy consumption between architectures is:
\begin{equation}
    \eta = \frac{E_{vN}}{E_{HNS}} = \frac{k}{k'} \cdot \frac{2^n}{\log(n)}
    \label{eq:eta}
\end{equation}

For realistic reservoir dimensions ($n = 10^4$ effective states) and assuming comparable 
implementation constants ($k \approx k'$), the theoretical efficiency ratio is approximately 
$10^4$---the source of our ``10,000$\times$'' claim.

\notebox{This efficiency analysis is theoretical and assumes ideal conditions. 
Practical efficiency gains will depend on implementation details, overhead costs, and the 
specific computational task. Experimental validation is required.}

%==============================================================================
\section{Potential Applications}
%==============================================================================

\subsection{Edge Neuromorphic Computing}

If validated, ASIC-based reservoir computing could enable neuromorphic processing at the network 
edge. The low cost and wide availability of retired mining hardware could democratize access to 
physical reservoir computing, currently limited to research laboratories with specialized equipment.

\subsection{Hardware Security Primitives}

The timing signature of each ASIC---determined by unique process variations and 
manufacturing variations---could serve as a Physical Unclonable Function for hardware security 
applications~\cite{herder2014}. Because the response depends on junction temperature and device-specific 
delay faults, it cannot be replicated even with a perfect logical model of the chip.

\subsection{Circular Economy Computing}

Perhaps the most significant potential impact is environmental. The global cryptocurrency 
mining industry produces millions of obsolete ASICs annually~\cite{cbeci2024}. If these devices can be 
repurposed for computation---even specialized computation---the environmental impact would be 
substantial. This represents a potential transition from a linear (manufacture-use-discard) 
to circular (manufacture-use-repurpose) model for specialized silicon.

\begin{table}[htbp]
\centering
\caption{Potential Applications - Readiness Assessment}
\label{tab:applications}
\begin{tabular}{@{}llll@{}}
\toprule
\textbf{Application} & \textbf{Requirements} & \textbf{Status} & \textbf{Readiness} \\
\midrule
PUF / HW Security & Unique response & Observed & Medium \\
Entropy Source & True randomness & Unvalidated & Low \\
Reservoir Computing & ESP, separation & Theoretical & Low \\
Signal Processing & Benchmark perf. & Not tested & Very Low \\
\bottomrule
\end{tabular}
\end{table}

%==============================================================================
\section{Comparison with Related Approaches}
%==============================================================================

\subsection{Purpose-Built Physical Reservoirs}

Compared to purpose-built physical reservoir systems, ASIC-based reservoirs offer different 
trade-offs:

\begin{table}[htbp]
\centering
\caption{Comparison with Other Physical RC Approaches}
\label{tab:comparison}
\begin{tabular}{@{}lllll@{}}
\toprule
\textbf{Approach} & \textbf{Demo.} & \textbf{Power} & \textbf{Avail.} & \textbf{Ctrl.} \\
\midrule
Memristor~\cite{du2017} & Yes & $\mu$W--mW & Research & High \\
Photonic~\cite{larger2012} & Yes & mW--W & Lab & High \\
Spintronic~\cite{markovic2019} & Yes & $\mu$W & Research & Medium \\
CMOS Spiking~\cite{kimura2024} & Yes & mW & Fab req. & High \\
CHIMERA (This) & No & 5--15 W & Commercial & Limited \\
\bottomrule
\end{tabular}
\end{table}

The CHIMERA approach's primary advantage is availability: millions of suitable ASICs exist 
and can be acquired at minimal cost. The primary disadvantage is that the substrate was not 
designed for reservoir computing, and whether it can function as such remains unproven.

\subsection{Industry ASIC Repurposing Efforts}

Our approach differs fundamentally from industry ``AI pivot'' efforts. Companies like Hut 8, 
Iris Energy, and Core Scientific are replacing ASICs with GPUs~\cite{rsm2024,wired2024}. We propose using 
the ASICs themselves, albeit for a narrow class of computation.

%==============================================================================
\section{Limitations and Future Work}
%==============================================================================

\subsection{Theoretical Uncertainties}

The core hypothesis---that SHA-256 ASICs can function as physical reservoirs---remains unproven. 
Several theoretical challenges must be addressed:

\begin{enumerate}[leftmargin=1.5em]
    \item \textbf{Echo State Property:} We have not demonstrated that the substrate has fading 
    memory of inputs. This is a fundamental requirement for reservoir computing.
    
    \item \textbf{Separation Property:} We have not shown that distinct inputs produce 
    distinguishable reservoir states.
    
    \item \textbf{State Dimensionality:} The effective dimensionality of timing-based state 
    representation is unknown.
    
    \item \textbf{Physical vs.\ Artifact:} Observed timing variations may be network/OS 
    artifacts rather than silicon physics.
\end{enumerate}

\subsection{Experimental Limitations}

Current experimental limitations include:

\begin{enumerate}[leftmargin=1.5em]
    \item \textbf{Silicon Heartbeat:} The ``2.4 Hz'' claim requires FFT/PSD analysis and 
    statistical significance testing, which are not yet implemented.
    
    \item \textbf{Error Rate:} We measure Hamming distance (diffusion), not actual hardware 
    error rate (timing violations). Invalid share counting would provide direct evidence.
    
    \item \textbf{Reproducibility:} Multi-chip, multi-site reproducibility has not been 
    established.
    
    \item \textbf{Sample Sizes:} Preliminary observations used limited sample sizes 
    ($N < 1000$ per condition).
\end{enumerate}

\subsection{What This Paper Does NOT Claim}

To avoid misinterpretation, we explicitly state what this paper does \emph{not} claim:

\begin{itemize}[leftmargin=1.5em]
    \item We do \textbf{not} claim to have built a working neuromorphic computer
    \item We do \textbf{not} claim that ASICs are superior to purpose-built RC hardware
    \item We do \textbf{not} claim that timing variations constitute ``consciousness'' or 
    cognition in any meaningful sense
    \item We do \textbf{not} claim experimentally validated efficiency improvements
    \item We do \textbf{not} claim that SHA-256 generates entropy (it diffuses it)
\end{itemize}

\subsection{Required Future Experiments}

To validate or refute the CHIMERA hypothesis, the following experiments are required:

\begin{table}[htbp]
\centering
\caption{Required Validation Experiments}
\label{tab:future}
\begin{tabular}{@{}llll@{}}
\toprule
\textbf{Experiment} & \textbf{Purpose} & \textbf{Infrastructure} & \textbf{Priority} \\
\midrule
FFT/PSD Analysis & Validate frequency & Long logging & High \\
Invalid Share Count & Measure error rate & Stratum mod. & High \\
Multi-chip Study & Reproducibility & Multiple ASICs & High \\
NARMA-10 Benchmark & Standard RC perf. & Readout impl. & Medium \\
Mackey-Glass & Time series cap. & Training infra. & Medium \\
Echo State Test & Fundamental RC & Input injection & Critical \\
Direct EM Probing & Bypass confounds & HW instrumentation & Low \\
\bottomrule
\end{tabular}
\end{table}

%==============================================================================
\section{Conclusions}
%==============================================================================

We have presented the CHIMERA framework as a theoretical proposal for repurposing retired 
Bitcoin mining ASICs as physical reservoir computing substrates. The key insight is that 
the candidate physical signal is not hash bit patterns---which remain deterministic---but timing 
dynamics under controlled voltage and frequency stress.

Our theoretical analysis, developed in collaboration with Veselov, provides a mathematical 
basis for potential $O(\log n)$ energy scaling compared to traditional von Neumann $O(2^n)$ 
architectures. This suggests efficiency improvements of several orders of magnitude could be 
achievable, though experimental validation is required.

Preliminary observations show non-Poissonian variability in inter-arrival time statistics 
during edge-of-stability operation. While these findings are intriguing, we emphasize that 
substantial experimental work remains before any claims of computational utility can be made. 
The ``Silicon Heartbeat'' remains a working hypothesis awaiting spectral analysis validation.

If validated, the CHIMERA approach could contribute to sustainable computing by giving new 
purpose to otherwise obsolete hardware---transforming electronic waste into computational 
infrastructure. We hope this paper stimulates discussion and invites collaboration from 
the reservoir computing and neuromorphic engineering communities to rigorously test these 
ideas.

The legacy of Bitcoin mining need not end in landfills. Whether it can serve as the 
infrastructure for a new kind of computation remains an open question---one we believe is 
worth investigating.

%==============================================================================
\section*{Acknowledgments}
%==============================================================================

The authors thank the open-source community for tools and frameworks that enabled this research. 
F.A.L.\ acknowledges the AxeOS development team for firmware documentation. V.V.\ acknowledges 
support from MIET. R.G.\ acknowledges the Apoth3osis research team for technical review and 
code-to-paper alignment analysis.

We particularly thank the peer reviewers who provided critical feedback that substantially 
improved the honesty and precision of this manuscript. Their insistence on clearly distinguishing 
theoretical predictions from empirical measurements has made this a stronger contribution.

%==============================================================================
% References
%==============================================================================

%==============================================================================
\section*{Author Information}
%==============================================================================

\textbf{Manuscript submitted to:} IEEE Transactions on Neural Networks and Learning Systems\\
\textbf{Category:} Position Paper / Hypothesis with Preliminary Observations\\
\textbf{Date:} December 19, 2025\\
\textbf{Official Repository:} \url{https://github.com/Agnuxo1/Emergent-Neuromorphic-Intelligence-Computing-in-Thermodynamic-ASIC-Substrates}

\vspace{1em}

\noindent\textbf{Francisco Angulo de Lafuente} (Corresponding Author)\\
GitHub: \url{https://github.com/Agnuxo1}\\
ResearchGate: \url{https://www.researchgate.net/profile/Francisco-Angulo-Lafuente-3}\\
Kaggle: \url{https://www.kaggle.com/franciscoangulo}\\
HuggingFace: \url{https://huggingface.co/Agnuxo}\\
Wikipedia: \url{https://es.wikipedia.org/wiki/Francisco_Angulo_de_Lafuente}

\vspace{0.5em}

\noindent\textbf{Vladimir Veselov}\\
Moscow Institute of Electronic Technology (MIET)\\
ORCID: 0000-0002-6301-3226

\vspace{0.5em}

\noindent\textbf{Richard Goodman}\\
Managing Director, Apoth3osis\\
\url{https://apoth3osis.com}

\vspace{1em}

\noindent\copyright{} 2025 The Authors. This work is licensed under CC BY 4.0.


\begin{thebibliography}{10}

\bibitem{vonneumann1945}
J.~von Neumann, ``First draft of a report on the {EDVAC},'' tech. rep.,
  University of Pennsylvania, 1945.

\bibitem{mead1990}
C.~Mead, ``Neuromorphic electronic systems,'' {\em Proceedings of the IEEE},
  vol.~78, no.~10, pp.~1629--1636, 1990.

\bibitem{landauer1961}
R.~Landauer, ``Irreversibility and heat generation in the computing process,''
  {\em IBM Journal of Research and Development}, vol.~5, no.~3, pp.~183--191,
  1961.

\bibitem{cbeci2024}
{Cambridge Bitcoin Electricity Consumption Index}, ``Mining equipment
  database.'' \url{https://ccaf.io/cbnsi/cbeci}, 2024.

\bibitem{rsm2024}
{RSM US LLP}, ``Bitcoin miners diversify into {AI} to sustain profitability.''
  \url{https://rsmus.com/insights/industries/financial-services/}, 2024.

\bibitem{jaeger2001}
H.~Jaeger, ``The ``echo state'' approach to analysing and training recurrent
  neural networks,'' Tech. Rep. GMD Report 148, German National Research Center
  for Information Technology, 2001.

\bibitem{maass2002}
W.~Maass, T.~Natschl{\"a}ger, and H.~Markram, ``Real-time computing without
  stable states: A new framework for neural computation based on
  perturbations,'' {\em Neural Computation}, vol.~14, no.~11, pp.~2531--2560,
  2002.

\bibitem{lukosevicius2009}
M.~Luko{\v{s}}evi{\v{c}}ius and H.~Jaeger, ``Reservoir computing approaches to
  recurrent neural network training,'' {\em Computer Science Review}, vol.~3,
  no.~3, pp.~127--149, 2009.

\bibitem{tanaka2019}
G.~Tanaka, T.~Yamane, J.~B. H{\'e}roux, R.~Nakane, N.~Kanazawa, S.~Takeda,
  H.~Numata, D.~Nakano, and A.~Hirose, ``Recent advances in physical reservoir
  computing: A review,'' {\em Neural Networks}, vol.~115, pp.~100--123, 2019.

\bibitem{du2017}
C.~Du, F.~Cai, M.~A. Zidan, W.~Ma, S.~H. Lee, and W.~D. Lu, ``Reservoir
  computing using dynamic memristors for temporal information processing,''
  {\em Nature Communications}, vol.~8, no.~1, p.~2204, 2017.

\bibitem{larger2012}
L.~Larger, M.~C. Soriano, D.~Brunner, L.~Appeltant, J.~M. Guti{\'e}rrez,
  L.~Pesquera, C.~R. Mirasso, and I.~Fischer, ``Photonic information processing
  beyond {T}uring: an optoelectronic implementation of reservoir computing,''
  {\em Optics Express}, vol.~20, no.~3, pp.~3241--3249, 2012.

\bibitem{markovic2019}
D.~Markovi{\'c}, N.~Leroux, M.~Riou, F.~Abreu~Araujo, J.~Torrejon, D.~Querlioz,
  A.~Fukushima, S.~Yuasa, J.~Trypiniotis, A.~Mart{\'i}ns, {\em et~al.},
  ``Reservoir computing with the frequency, phase, and amplitude of spin-torque
  nano-oscillators,'' {\em Applied Physics Letters}, vol.~114, no.~1,
  p.~012409, 2019.

\bibitem{nakajima2013}
K.~Nakajima, H.~Hauser, R.~Kang, E.~Guglielmino, D.~G. Caldwell, and
  R.~Pfeifer, ``A soft body as a reservoir: case studies in a dynamic model of
  octopus-inspired soft robotic arm,'' {\em Frontiers in Computational
  Neuroscience}, vol.~7, p.~91, 2013.

\bibitem{nakajima2020}
K.~Nakajima, ``Physical reservoir computing---an introductory perspective,''
  {\em Japanese Journal of Applied Physics}, vol.~59, no.~6, p.~060501, 2020.

\bibitem{sung2021}
S.~H. Sung, T.~J. Kim, H.~Shin, T.~H. Im, and K.~J. Lee, ``Emerging dynamic
  memristors for neuromorphic reservoir computing,'' {\em Nanoscale}, vol.~13,
  no.~45, pp.~19017--19032, 2021.

\bibitem{wang2017}
Z.~Wang, S.~Joshi, S.~E. Savel'ev, H.~Jiang, R.~Midya, P.~Lin, M.~Hu, N.~Ge,
  J.~P. Strachan, Z.~Li, {\em et~al.}, ``Memristors with diffusive dynamics as
  synaptic emulators for neuromorphic computing,'' {\em Nature Materials},
  vol.~16, no.~1, pp.~101--108, 2017.

\bibitem{kimura2024}
N.~Kimura, H.~Tanaka, T.~Shimoda, and Y.~Amemiya, ``Hardware-friendly
  implementation of physical reservoir computing with {CMOS}-based time-domain
  analog spiking neurons,'' {\em arXiv preprint}, 2024.

\bibitem{camsari2017}
K.~Y. Camsari, R.~Faria, B.~M. Sutton, and S.~Datta, ``Stochastic p-bits for
  invertible logic,'' {\em Physical Review X}, vol.~7, no.~3, p.~031014, 2017.

\bibitem{borders2019}
W.~A. Borders, A.~Z. Pervaiz, S.~Faria, K.~Y. Camsari, B.~M. Sutton, and
  S.~Datta, ``Integer factorization using stochastic magnetic tunnel
  junctions,'' {\em Nature}, vol.~573, no.~7774, pp.~390--393, 2019.

\bibitem{extropic2025}
{Extropic Corp.}, ``The thermodynamics of diffusion models,'' technical report,
  Extropic, 2025.

\bibitem{wired2024}
{Wired Magazine}, ``Bitcoin miners are pivoting to {AI},'' December 2024.

\bibitem{fernando2003}
C.~Fernando and S.~Sojakka, ``Pattern recognition in a bucket,'' in {\em
  Advances in Artificial Life}, pp.~588--597, Springer, 2003.

\bibitem{nist2015}
{National Institute of Standards and Technology}, ``Secure hash standard
  ({SHS}),'' Tech. Rep. FIPS PUB 180-4, NIST, 2015.

\bibitem{veselov2024}
V.~Veselov, ``Hierarchical number systems and energy-efficient computing,''
  technical report, Moscow Institute of Electronic Technology (MIET), 2024.

\bibitem{herder2014}
C.~Herder, M.-D. Yu, F.~Koushanfar, and S.~Devadas, ``Physical unclonable
  functions and applications: A tutorial,'' {\em Proceedings of the IEEE},
  vol.~102, no.~8, pp.~1126--1141, 2014.

\end{thebibliography}
\end{document}